\documentclass[runningheads]{llncs}
\usepackage{graphicx}

\usepackage{tikz}
\usepackage{comment}
\usepackage{amsmath,amssymb} 
\usepackage{color}
\usepackage{multirow}

\usepackage[accsupp]{axessibility}  

\usepackage{amssymb}
\usepackage{pifont}
\newcommand{\cmark}{\ding{51}}%
\newcommand{\xmark}{\ding{55}}%

\begin{document}
\pagestyle{headings}
\mainmatter
\def\ECCVSubNumber{28}  

\title{Is GPT-3 all you need for Visual Question Answering in Cultural Heritage?} 

\titlerunning{Is GPT-3 all you need for Visual Question Answering in Cultural Heritage?}
%
\author{Pietro Bongini\orcidID{0000-0002-8217-6266} \and
Federico Becattini\orcidID{0000-0003-2537-2700} \and
Alberto Del Bimbo\orcidID{0000-0002-1052-8322}}
\authorrunning{P. Bongini et al.}
%
\institute{University of Florence, MICC, Italy\\
\email{name.surname@unifi.it}}
\maketitle

\begin{abstract}
The use of Deep Learning and Computer Vision in the Cultural Heritage domain is becoming highly relevant in the last few years with lots of applications about audio smart guides, interactive museums and augmented reality. All these technologies require lots of data to work effectively and be useful for the user. In the context of artworks, such data is annotated by experts in an expensive and time consuming process. In particular, for each artwork, an image of the artwork and a description sheet have to be collected in order to perform common tasks like Visual Question Answering. In this paper we propose a method for Visual Question Answering that allows to generate at runtime a description sheet that can be used for answering both visual and contextual questions about the artwork, avoiding completely the image and the annotation process. For this purpose, we investigate on the use of GPT-3 for generating descriptions for artworks analyzing the quality of generated descriptions through captioning metrics. Finally we evaluate the performance for Visual Question Answering and captioning tasks.

\end{abstract}

\section{Introduction}
Cultural Heritage often relies on digital resources to engage and attract visitors. From audio-guides to smartphone applications, museum visits are becoming increasingly more interactive, allowing users to deepen concepts without the need of a human assistant or after the visit is concluded.
Forms of gamification are also important, favoring engagement especially for young visitors and instructional purposes.
Artificial Intelligence and Computer Vision are playing a large part in the development of such smart visits and applications~\cite{fiorucci2020machine,becattini2016imaging,bongini2020visual,cetinic2022understanding}. A notable machine learning application that has recently found usage in cultural heritage is Visual Question Answering (VQA), which exploits both Computer Vision and Natural Language Processing to allow users to ask questions on the content of an image~\cite{bongini2020visual}.
The advantage of VQA is that it allows museums to develop smart guides and interactive gamification approaches. However, for pictorial art, most questions posed by users concern contextual information rather than what is actually depicted in a painting.

To address this limitation, an evolution of VQA known as Contextual Question Answering (CQA) was proposed~\cite{bongini2020visual}. The authors explicitly focused on cultural heritage applications, combining visual and contextual cues to answer questions.
The contextual information is derived from a textual meta-data, which is fed to the model along with the question and the image. In this way the VQA/CQA model has to learn to attend either relevant parts of an image or relevant sections of the text to provide an adequate answer.
The need of a textual data nonetheless opens a new issue, namely where to obtain such description. Information sheets for artworks may already be available to museum curators yet extending this kind of application to new data becomes time-consuming and requires a domain expert.

In this paper we explore the usage of a generative natural language processing model to automatically create contextual information to be fed to a CQA model.
In fact, recently, generative text models have been finding large diffusion with groundbreaking results. Among these we find GPT-3, a generative model trained on a massive corpus of textual data regarding several domains, including art \cite{brown2020language}.
GPT-3 is capable of generating a description starting from a textual query and it has been demonstrated that the model includes knowledge of the entities described in the training data, for example paintings and artworks.
We therefore investigate the possibilities and the limitations of GPT-3 in applications for cultural heritage, with a specific focus on question answering.
In particular, we explore the quality of the textual description of artworks that the model is able to generate and we evaluate their applicability for visual and contextual question answering.

The main contributions of our work are the following:
\begin{itemize}
    \item We propose an automatic approach to generate textual information sheets of artworks exploiting GPT-3. We find that the model has excellent knowledge of art concepts and event details of specific paintings.
    \item We propose a method to answer both visual and contextual questions which is artwork agnostic, i.e. it does not require any additional data or training to be adapted to a new set of images.
    \item We explore the applicability of GPT-3 in cultural heritage applications. To the best of our knowledge we are the first to apply GPT-3 to the art domain.
\end{itemize}

\section{Related Work}

Natural Language Processing (NLP) in recent years has evolved at an extremely fast pace, converging to a set of well defined application 
paradigms~\cite{sun2022paradigm}. Such paradigms include text classification, matching, machine reading comprehension, sequence to sequence translation, sequence tagging and language modeling.
Despite the wide variety of tasks \cite{seidenari2021language,anderson2018bottom,cornia2020meshed}, some recent noticeable approaches have been shown to perform well as generic pre-training for NLP models \cite{devlin2018bert,brown2020language}. In particular, this can be attributed to the introduction of attention models, based on the transformer architecture \cite{vaswani2017attention}. The effectiveness of models such as BERT \cite{devlin2018bert} stems from the capability of processing text bidirectionally exploiting the self-attention mechanism of transformers to obtain word level representations that are informed of their surrounding context within the sentence. Whereas BERT is built exploiting the encoder part of the transformers, another state of the art approach for NLP, Generative Pre-trained Transformer (GPT) \cite{radford2018improving}, is built stacking transformer decoder blocks and is trained to predict the next word in a sentence.
The model has then been improved in subsequent versions, GPT-2 \cite{radford2019language} and GPT-3 \cite{brown2020language}, yielding larger and more effective models. 

Interestingly, GPT-3 has been trained using a large quantity of internet data, meaning that the training process has distilled into the model common sense knowledge making it able to generate essays and even poetry \cite{dale2021gpt}.
In this paper we exploit GPT-3 as a generator of textual content describing artworks, showing that it can be used for interactive applications for cultural heritage such as captioning~\cite{liu2017improved} and Visual Question Answering (VQA) \cite{antol2015vqa}.
VQA is a recent trend in machine learning that bridges the Natural Language Processing and Computer Vision domains \cite{barra2021visual}. The goal is to answer questions regarding the content of an image through artificial intelligence. This involves several sub-tasks such as object detection~\cite{han2018advanced} and recognition~\cite{kheradpisheh2018stdp}, question reasoning~\cite{lu2016hierarchical}.
Typical VQA approaches use Convolutional Neural Networks (CNNs) to interpret images and Recurrent Neural Networks (RNNs) to process questions. The authors of ~\cite{anderson2018bottom} proposed a bottom-up attention mechanism looking at salient objects in images. Differently from previous approaches that considered regularly spaced image portions ~\cite{shih2016look}, they use object Faster R-CNN \cite{ren2015faster} features as attention candidates. In the past few years multiple Transformer-based approaches reached impressive performances on this task ~\cite{su2019vl,tan2019lxmert,li2020oscar,wang2022unifying}.

Recently, a few approaches \cite{bongini2020visual,garcia2020dataset,Vannoni2020DataCF,asprino2021large} have addressed VQA in the cultural heritage domain.
A dataset of questions and answers for art related questions has been recently proposed \cite{asprino2021large}, exploiting an ontology based framework to extract data with question templates.
The authors of \cite{bongini2020visual} and \cite{garcia2020dataset} found that to make the best out of VQA for museum applications, a model must be able to integrate some source of external knowledge in order to address contextual questions, i.e. questions concerning non-visual cues such as name of the author, year and artistic style. In particular, \cite{bongini2020visual} used a question classifier to understand if visual of contextual knowledge is required. Depending on the output of the classifier a VQA model is used, otherwise a purely textual based question answering model is used discarding the image content.
In this work we explore the effectiveness of using GPT-3 to generate artwork captions, suitable for such a visual and contextual question answering model.

Other approaches have been used to answer questions relying on captions, yet only regarding visual content \cite{sheng2019can}. The most similar approach to ours is instead \cite{yang2022empirical}, which used GPT-3 for VQA. However, differently from us, the authors feed GPT-3 with questions and descriptions generated by an image captioner directly to obtain an answer. We, instead, aim at extracting the domain specific knowledge from GPT-3 which is requested to correctly answer a question.

\section{GPT-3}

To provide to the reader a better understanding of our work, here we present a brief background context about GPT-3, the third version of Generative Pre-Trained Transformer \cite{brown2020language}. This is an autoregressive language model with 175 billion parameters that can be used for different tasks without any finetuning, achieving strong performances.

The architecture of the GPT-3 Transfomer model is made of 96 attention layers. While language models like BERT \cite{devlin2018bert} use the Encoder to generate embeddings from the raw text which can be used in other machine learning applications, GPT-3 use the Decoder half, so it takes embeddings as inputs and produces text. In particular the GPT-3 language model has the ability to generate natural language text that can be hard to distinguish from human-written text, to the point that research has been carried out to asses whether GPT-3 could pass a written Turing test \cite{elkins2020can}.


Concretely, during inference, the target of the new task $y$ is directly predicted conditioned on the given context $C$ and the new task’s input $x$, as a text sequence generation task. Note that all $C$, $x$ and $y$ are text sequences. For example, $y = (y^1,..., y^T)$.
Therefore, at each decoding step $t$ we have 
\begin{equation}
    y^t= \arg \max_{y^t}{p_{W}}(y^t|C,x, y<t)
\end{equation}
where $W$ are the weights of the pretrained language model, which are frozen for all new tasks. The context
$C={h,x_1,y_1,...,x_n,y_n}$ consists of an optional prompt head $h$ and $n$ in-context examples $(\{x_i,y_i\}{^n}_{i=1})$ from the new task.

\section{Method}
\label{ref:method}
\begin{figure}[t]
\centering
\includegraphics[width=\textwidth]{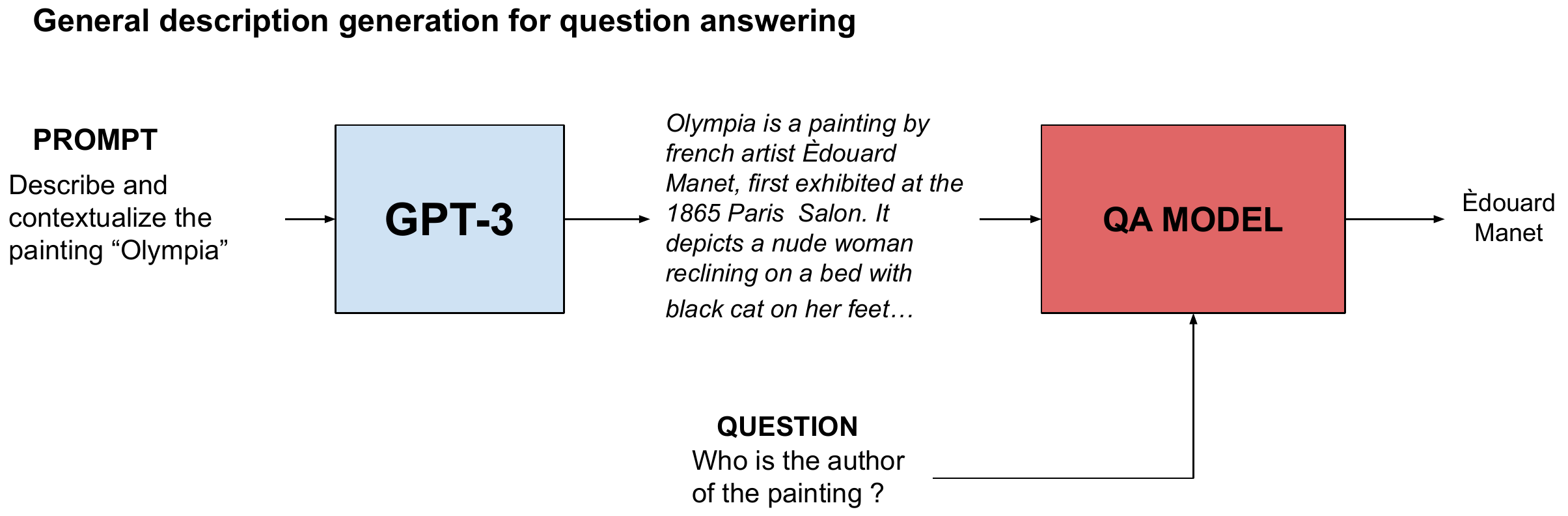}
\caption{Scheme of our method for answering questions using a general generated description. A prompt with a specific structure is given in input to GPT-3. Subsequently the generated text is fed together with the question to a Question Answering model that outputs the answer.}
\label{fig:general}
\end{figure}

\begin{figure}[t]
\centering
\includegraphics[width=\textwidth]{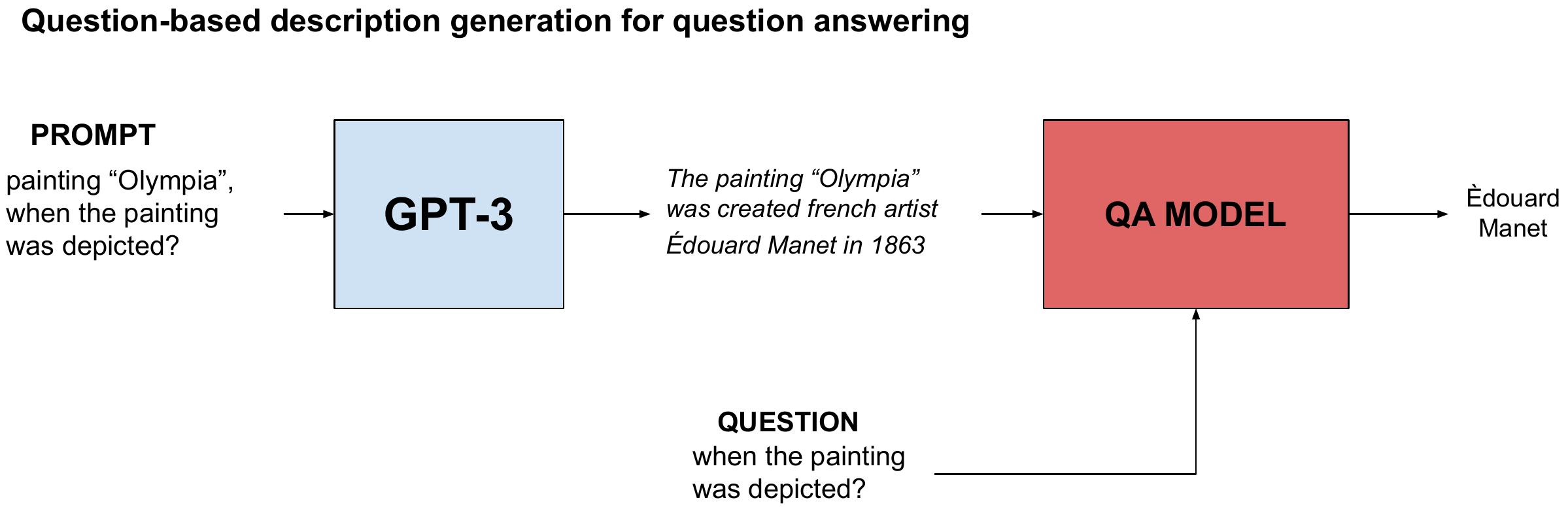}
\caption{Scheme of our method for answering questions using a question-based generated description. A prompt containing the name of the painting and the question is given in input to GPT-3. Subsequently the generated text is fed together with the question to a Question Answering model that outputs the answer.}
\label{fig:question_based}
\end{figure}

In a Cultural Heritage context, the information useful to answer questions about a specific artwork is contained in the artwork image and in its contextual description.
Finding such a description might not be trivial, since it might require a domain expert to write it down. At the same time, it is quite costly to train a Visual Question Answering model that takes in input both the image and the description. This is also not straightforward, since the two modalities need to be blended and matched together.
Consequently, the main idea of this work is to generate new descriptions for artworks based on a specific prompt or a specific question and directly use these descriptions to answer visual and contextual questions.
The overall pipeline of our proposed work is as follows:
\begin{enumerate}
    \item \textbf{GPT-3 caption generation.} We use GPT-3 to generate descriptions of artworks, leveraging its memorization capabilities that allowed the model retain relevant information about training instances.
    An important aspect in this phase in to feed the correct prompt in input to GPT-3 in order to obtain realistic and correct descriptions. We consider two different types of input prompt: \\
    
    \begin{itemize}
        \item \textbf{General} - A general prompt where the expected output is a general description of the artwork. The input text follows the structure:
        
        \texttt{"Describe and Contextualize the painting $<$ painting\_name $>$"}\\
        
        \item \textbf{Question-based} - A specific question based prompt. The input text follows the structure:
        
        \texttt{"Painting $<$ painting\_name $>$ $<$ question $>$"}.
        
        The expected generated text by GPT-3 is a small text snippet that consists in a couple of sentences, focused on the topic of the question. \\
    \end{itemize}
    
    \item \textbf{Question answering.} Once the description has been generated in the previous step, we can exploit it to answer both visual and contextual questions through a Question Answering language model. For this purpose we use a pretrained version of DistilBert~\cite{sanh2019distilbert} fine-tuned on the SQUAD~\cite{rajpurkar2016squad} dataset. We feed in input to the DistilBert model the generated text from the previous step together with the question. The answer given as output will be the final answer of our method.
\end{enumerate}
Fig.~\ref{fig:general} and Fig.~\ref{fig:question_based} show a scheme of the two variants of our method. More precisely, in Fig.~\ref{fig:general} the general input prompt for GPT-3 yields the generation of a long description of the artwork (similar to a museum information sheet).
On the other hand, the question-based prompt in Fig.~\ref{fig:question_based} yields only the generation of a brief output text, which we find suitable for answering the question.
In conclusion, these two schemes follow roughly the same structure. The difference is in the input prompt that in the case of Fig.~\ref{fig:general} is more general and in Fig.~\ref{fig:question_based} is more task oriented.
    
\section{Experiments}
In this section we first outline the experimental setting for the experiments carried out in this paper, presenting dataset and experimental protocol and we then move on to a discussion of the results.

\subsection{Dataset}
For our experiments, we use the Artpedia dataset \cite{stefanini2019artpedia}. Artpedia contains a collection of 2,930 artworks, associated to a variable number of textual descriptions gathered from WikiPedia.
Sentences are labelled as a visual descriptions or as a contextual descriptions. Contextual descriptions regard information about the artwork that does not directly describe its visual content.
For instance, contextual descriptions can describe the historical context of the artwork, its author, the artistic influence or the museum where a painting is exhibited.
The dataset contains 28,212 descriptions, 9,173 of which are labelled as visual and the remaining 19,039 as contextual.
The Artpedia dataset has been extended with Question-Answer annotations in \cite{bongini2020visual}. In fact, a subset of the images have been associated with visual and contextual questions, derived from the corresponding captions. In this work we follow the dataset split of \cite{bongini2020visual}.

\subsection{Experimental protocol}
Following prior work such as \cite{bongini2020visual}, we evaluate visual questions and contextual questions with different metrics. In fact, visual question answering and traditional text-based question answering are often treated in two different ways. Visual Question Answering is considered as a classification problem, meaning that a model has to pick an answer from a predefined dictionary of possible candidates containing a few words each. This stems from the fact that questions in most datasets are a way of guiding attention towards specific objects or attributes in the image, without requiring any complex form of language reasoning. Question Answering on the other hand is based on a set of sentences, which may contain rare or out-of-dictionary words. The task is in fact defined as identifying a subset of the textual description that contains the answer.

In light of this, to evaluate visual questions we rely on accuracy:
\begin{equation}\label{eq:1}
 Accuracy = \frac{N_c}{N_a}
\end{equation}{}
where $N_c$ is the number of correct answers and $N_a$ the number of total answers.

For text-based question answering, instead, we use both accuracy and F1-measure, a metric that takes into account the global correctness of the answer:

\begin{equation}
    F1 = 2 \times \frac{Precision \times Recall}{Precision + Recall}
\end{equation}{}

Where $Precision$ is defined as:

\begin{equation}
   Precision = \frac{N_{Cw}}{|ans|}
\end{equation}{}

with $N_{Cw}$ is the number of common words between the output answer and the ground truth answer and $ans$ the number of words in the generated answer.

$Recall$ instead is defined as:

\begin{equation}
   Recall = \frac{N_{Cw}}{|gt|}
\end{equation}{}

where $|gt|$ is the number of words in the ground truth.

We also evaluate the quality of the descriptions generated by GPT-3, considering it as a standalone image captioning model. We use the following standard metrics for captioning:
\begin{itemize}
    \item \textit{BLEU1} \cite{papineni2002bleu}: BiLingual Evaluation Understudy (BLEU) is the most commonly used metric for machine translation and image captioning. BLEU scores are based on how similar a generated caption is to a reference caption, computing the precision of the generated words. The downside of BLEU is that it is very sensitive to small changes, such as synonyms or different word order.
    \item \textit{ROUGE} \cite{lin2004rouge}: differently from BLEU, which measures the precision of the caption, Recall Oriented Understudy of Gisting Evaluation (ROUGE) focuses on quantifying the amount of correct words with respect to the reference. Thus, this metric is recall-based and tends to reward long sentences.
    \item \textit{CIDEr} \cite{vedantam2015cider}: Consensus-based Image Description Evaluation (CIDEr) is an automatic consensus metric that measures the similarity of captions against a set of ground truth sentences written by humans. This metric has been shown to yield a higher agreement with humans generated text since it captures notions of grammar, importance and precision and recall.
    \item \textit{Cosine Similarity}: we compute the cosine similarity between feature vectors for the generated caption and the reference caption. Features are extracted with the algorithm TF-IDF \cite{salton1988term}.
\end{itemize}

\subsection{Experimental Results}

\begin{table}[t]
\caption{Image captioning results. We compare our method which generates captions with GPT-3 with the \textit{General} and the \textit{Question-based} approaches. In the \textit{Question-based} approach we concatenate all the outputs of GPT-3 after conditioning it with different questions related to the image. We compare the results against visual captions, contextual captions or both.}
\label{tab:caption}
\centering
\begin{tabular}{l|c|ccc}
Description type~~~~ & ~~~~Metric~~~~ & ~~OFA \cite{wang2022unifying}~~   & ~~ Ours General~~ & Ours Question-based \\ \hline
\multirow{4}{*}{Visual}              & BLEU1                      & 0.048 & \textbf{0.181}        & 0.137               \\
                                     & ROUGE                      & 0.138 & \textbf{0.188}        & 0.16                \\
                                     & CIDEr                      & 0.091 & 0.079        & \textbf{0.172}               \\
                                     & COSINE                     & 0.113 & \textbf{0.157}        & 0.110                \\ \hline
\multirow{4}{*}{Contextual}          & BLEU1                      & 0.002 & \textbf{0.168}        & 0.160                \\
                                     & ROUGE                      & 0.062 & 0.178        & \textbf{0.179}               \\
                                     & CIDEr                      & 0.000     & \textbf{0.248}        & 0.129               \\
                                     & COSINE                     & 0.082 & 0.218        & \textbf{0.324}               \\ \hline
\multirow{4}{*}{All}                 & BLEU1                      & 0.000     & 0.113        & \textbf{0.185}               \\
                                     & ROUGE                      & 0.053 & 0.158        & \textbf{0.184}               \\
                                     & CIDEr                      & 0.000     & 0.016        & \textbf{0.098}               \\
                                     & COSINE                     & 0.122 & 0.253        & \textbf{0.341}              
\end{tabular}
\end{table}

\subsubsection{Captioning results}
We start by assessing the quality of the captions generated by GPT-3.
First of all, we ask GPT-3 to generate captions with our \textit{General} approach. In Tab. \ref{tab:caption} we compare the captions using as reference visual captions, contextual captions and both. All reference captions are ground truth captions taken from the Artpedia dataset \cite{stefanini2019artpedia}.

Interestingly, the model appears to better results for visual captions using BLEU1 and ROUGE metrics, while using CIDEr and cosine similarity, the model obtaines higher results for contextual captions. This may seem counter-intuitive but can be explained looking at the nature of the metrics. BLEU1 and ROUGE in fact respectively check for word-wise precision and recall, while CIDEr and cosine distance perform a sentence level scoring, which is closer to human consensus.
We observe that the model is able to obtain good results, especially with the cosine metric, even when using all the captions as reference.

We then evaluate the method by taking a concatenation of the outputs generated by GPT-3 after being conditioned by different questions related to the image.
This obviously introduces a strong bias, given also the fact that questions have been generated from information contained in the captions, but at the same time proves the usefulness of such captions for more advanced applications such as visual question answering. As can be seen in Tab. \ref{tab:caption}, conditioning GPT-3 with the captions leads to better captions according to most metrics.

In Tab. \ref{tab:caption} we also provide a baseline as reference, i.e. the output of the state of the art OFA captioning model \cite{wang2022unifying}. We observe that captions generated by OFA do not align well with the ground truth sentences. We attribute this to a domain shift between the datasets commonly used to train captioning models and descriptions of artworks. In fact, the former are sentences written by non-experts while for applications in cultural heritage a domain knowledge is required. This further motivates the usage of GPT-3, which seems to have integrated sufficient knowledge to articulate complex sentences with a domain specific jargon.

\begin{table}[t]
\centering
\caption{Experimental results for Visual Question Answering. We compare our approach against VQA-CH \cite{bongini2020visual} to understand whether GPT-3 can replace information sheets for artworks either for visual or contextual questions. We compare two versions of our model, the \textit{General} version, which produces generic descriptions of artworks and the \textit{Question-based} version, where prompts are conditioned with the input question to generate more specific descriptions.}
\label{tab:vqa}
\begin{tabular}{l|c|c|c|c}
 & Visual               & Contextual    & Accuracy & F1 score \\ \hline
VQA-CH \cite{bongini2020visual} & \xmark      & \cmark        & 0.684      & 0.832    \\
VQA-CH \cite{bongini2020visual} & \cmark     & \xmark         & 0.176      & 0.150     \\
VQA-CH \cite{bongini2020visual} & \cmark     & \cmark        & 0.504      & 0.417    \\ \hline
Ours - General          & \xmark              & \cmark        & 0.557      & 0.719    \\
Ours - General          & \cmark             & \xmark         & 0.070      & 0.055    \\
Ours - General          & \cmark             & \cmark        & 0.239      & 0.360     \\ \hline
Ours - Question-based   & \xmark              & \cmark        & 0.473      & 0.602    \\
Ours - Question-based   & \cmark             & \xmark         & 0.134      & 0.202    \\
Ours - Question-based   & \cmark             & \cmark        & 0.256      & 0.330    
\end{tabular}
\end{table}

\subsubsection{VQA results}
To evaluate the Visual Question Answering capabilities of our proposed method, we follow the setting of \cite{bongini2020visual}. However, we do not rely on any vision-based model but rather on a fully textual question answering model based on DistilBert \cite{sanh2019distilbert}, as explained in Sec. \ref{ref:method}.
In Tab. \ref{tab:vqa}, we compare our approach to the one of VQA-CH \cite{bongini2020visual}. It has to be noted that, contrary to \cite{bongini2020visual}, we do not rely on real textual descriptions, which are known to contain the answer, but we only extract information from GPT-3. This is a strong disadvantage for our method. However, we are not interested in obtaining better results than VQA-CH, but rather our goal is to demonstrate if GPT-3 can act as a substitute of textual descriptions handcrafted by domain experts.

We test our method evaluating the accuracy for visual questions, contextual questions and both together.
Quantitative results indicate that captions generated by GPT-3 can yield to high results for contextual questions, yet very low accuracy for visual questions.
As for the captioning setting, we impute this behavior to the fact that GPT-3 generates generic descriptions, without including a fine-grained description of the visual content.
Thus, on the one hand the question answering model is capable of extracting meaningful information from the generated captions. This means that GPT-3 is indeed capable of integrating domain knowledge during training and is capable of generating a complete information sheet of the artwork. On the other hand, captions appear to be too generic to obtain information about specific details in the image.

To overcome this limitation, we test the model using captions generated with out \textit{Question-based} approach. By feeding the answer to GPT-3 along with the title of the artwork, the model is able to generate more specific captions. Such captions, as explained in Sec. \ref{sec:qualitative} are usually shorter but are focused on the prompt. This is particularly interesting since it means that a purely text-based model is capable of addressing a vision-based task. In Tab. \ref{tab:vqa} it can be seen that for visual questions alone, our method with question-based captions performs on par or better than the vision-based VQA-CH model.

\begin{figure}[t]
\centering
\includegraphics[width=\textwidth]{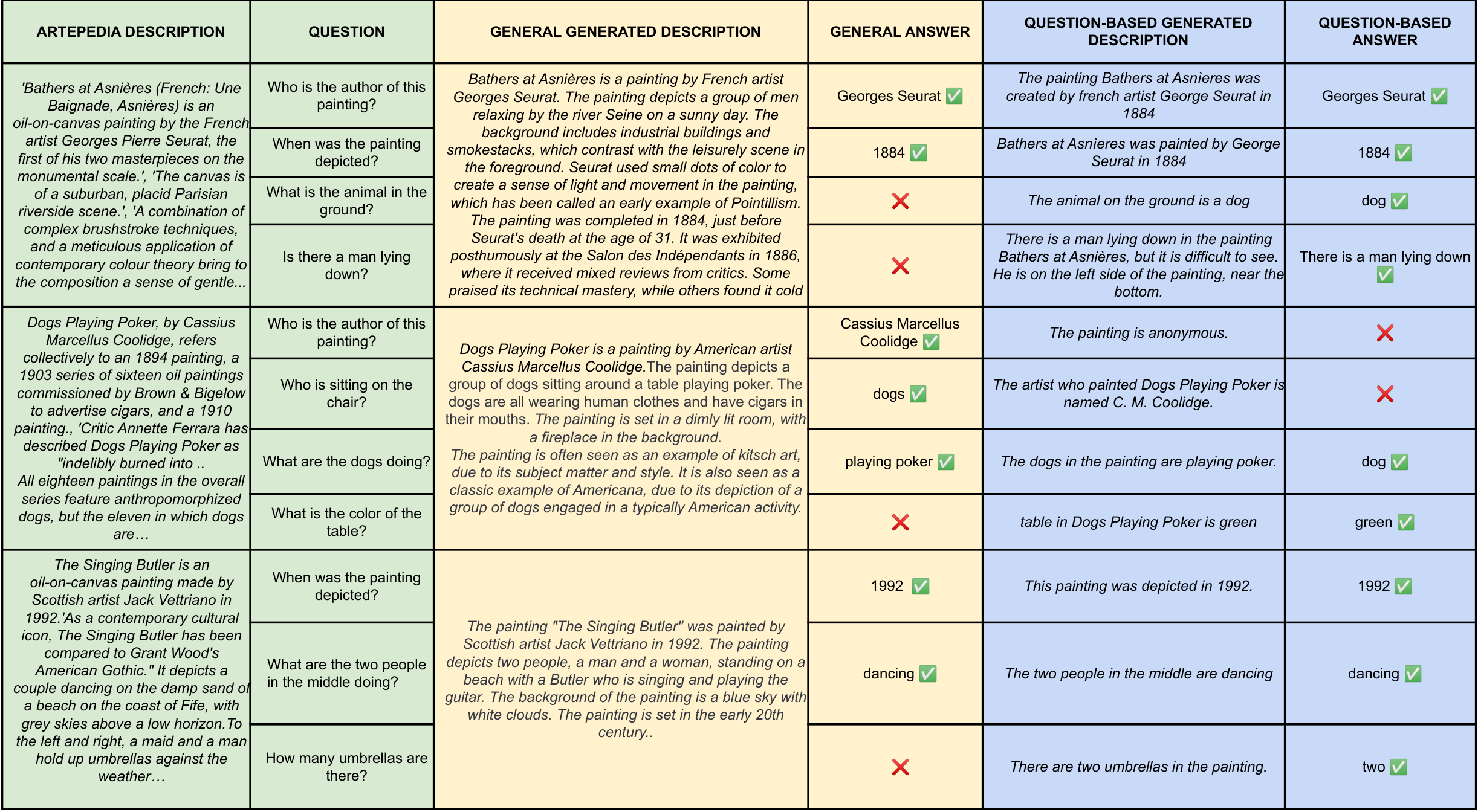}
\caption{Qualitative results of our method. \textit{Green}: ground truth description from the Artpedia dataset \cite{stefanini2019artpedia} and input question. \textit{Yellow}: general descriptions provided by GPT-3 and answer obtained based on such text. \textit{Blue}: Question-based description and correspondent answer.
General descriptions are longer and more detailed than question-based generated descriptions. However, question-based generated descriptions are customized for the specific question.}
\label{fig:qualitative_analysis}
\end{figure}

\section{Qualitative Analysis}
\label{sec:qualitative}
In this section we provide a qualitative analysis of the captions generated by GPT-3 in order to characterize which kind of information they contain in both the \textit{General} and \textit{Question-based} formulation.

Since the prompts that we feed to GPT-3 are different, with one being more general and the other being question-based, we expect that the corresponding generated text by GPT-3 will be different.
In Fig. \ref{fig:qualitative_analysis} we can observe these differences.
Generated general descriptions are very long and have the aspect of artwork information sheets in which we can find some visual and contextual information. Question-based generated descriptions are instead shorter and contain the knowledge needed to answer to the specific questions.
From Fig \ref{fig:qualitative_analysis} we can observe that the general description is very useful to answer to contextual questions but fails on some visual questions. This is likely due to different reasons:
\begin{itemize}
    \item The generated text does not take into account any specific question and this can lead to the generation of a description without specific information useful to answer to the question.
    \item Visual questions are very specific since they refer to object relationships, colors, counting, etc. and the GPT-3 model tends to be more shallow in generating its descriptions. 
\end{itemize}

On the other hand, question-based generated descriptions are helpful to answer visual questions but the small generated description useful to answer those specific questions could contain incorrect information leading to wrong answer predictions.
In conclusion these two ways of generating text to answer visual and contextual questions have some pros and cons:
\begin{itemize}
    \item General descriptions are longer and contain several pieces of information about the artwork. However this is fixed and could not contain the information needed to answer some questions.
    \item Question-based descriptions are generated for specific questions and contain only the information needed to answer the question on which GPT-3 has been conditioned. If the model has not memorized any specific information regarding such questions it may contain mistakes and descriptions will have to be re-computed for each question.
\end{itemize}

\section{Considerations on complexity and accessibility of GPT-3}
In the previous sections we have demonstrated that GPT-3 could indeed replace the usage of an information sheet handcrafted by a domain expert. However, we need to understand the actual applicability of GPT-3 in a real case application.
GPT-3 has 175B parameters, which approximately amounts to 700GB. This means that inference on a single GPU is unfeasible due to current technological limits. The model however has been made available from OpenAI and is accessible through API that have a pricing fee per generated token. These considerations somewhat limit a large-scale usage of the model, especially if a description has to be generated for each question to be answered. On the other hand, generating fixed descriptions offline, one for each artwork, appears a viable solution at least for addressing contextual questions.


\section{Conclusions}
In this paper we presented a method for Visual Question Answering in the Cultural Heritage domain. In particular we have addressed the problem of data annotation for artworks, generating descriptions with GPT-3. The performances for the VQA task show that the generated descriptions are useful to answer the questions correctly. This technique allows to answer visual and contextual questions focusing only on the generated description and can be used for any artwork. In fact, there is no need to retrain the model to incorporate new knowledge. This is possibile thanks to the memorization capabilities of GPT-3, which at training time has observed millions of tokens regarding domain-specific knowledge. Finally the generated description can be integrated as textual input (textual description) in a more complex architecture as \cite{bongini2020visual} in order to address tasks like Visual Question Answering. This is of particular interest for Cultural Heritage due to the domain shift between common VQA and captioning datasets compared to the technical jargon that is needed to properly address questions about art.

\paragraph{\textbf{Acknowledgements}}
This work was partially supported by the project ARS01 00421:
“PON IDEHA - Innovazioni per l’elaborazione dei dati nel settore del Patrimonio Culturale.”
This work was partially supported by the
European Commission under European Horizon 2020 Programme, grant number 101004545 - ReInHerit.

%
%
\bibliographystyle{splncs04}
\bibliography{egbib}
\end{document}